\documentclass[conference]{IEEEtran}
\IEEEoverridecommandlockouts
\usepackage{cite}
\usepackage{amsmath,amssymb,amsfonts}
\usepackage{algorithm}
\usepackage{algorithmicx}
\usepackage{algpseudocode}
\usepackage{graphicx}
\usepackage{textcomp}
\usepackage{xcolor}
\usepackage{tabularx}
\usepackage{titlesec}
    \titlespacing{\section}{3pt}{\parskip}{\parskip}
    \titlespacing{\subsection}{3pt}{\parskip}{\parskip}
\usepackage{enumitem}
\def\BibTeX{{\rm B\kern-.05em{\sc i\kern-.025em b}\kern-.08em
    T\kern-.1667em\lower.7ex\hbox{E}\kern-.125emX}}

\begin{document}

\title{Device-aware Optical Adversarial Attack for a Portable Projector-camera System
}
\author{\IEEEauthorblockN{Ning Jiang\IEEEauthorrefmark{1}\IEEEauthorrefmark{2},
Yanhong Liu\IEEEauthorrefmark{2},
Dingheng Zeng\IEEEauthorrefmark{2}, 
Yue Feng\IEEEauthorrefmark{2}, 
Weihong Deng\IEEEauthorrefmark{2}, 
and Ying Li\IEEEauthorrefmark{1}
}
\IEEEauthorblockA{\IEEEauthorrefmark{1}School of Software \& Microelectronics, Peking University, Beijing, China}
\IEEEauthorblockA{\IEEEauthorrefmark{2}Mashang Consumer Finance Co., Ltd., Chongqing, China}%
\IEEEauthorblockA{jiang.ning@stu.pku.edu.cn, \{yanhong.liu,dingheng.zeng,yue.feng,weihong.deng\}@mxsf.com, li.ying@pku.edu.cn} 
}

\maketitle

\begin{abstract}
Deep-learning-based face recognition (FR) systems are susceptible to adversarial examples in both digital and physical domains. Physical attacks present a greater threat to deployed systems as adversaries can easily access the input channel, allowing them to provide malicious inputs to impersonate a victim. This paper addresses the limitations of existing projector-camera-based adversarial light attacks in practical FR setups. By incorporating device-aware adaptations into the digital attack algorithm, such as resolution-aware and color-aware adjustments, we mitigate the degradation from digital to physical domains. Experimental validation showcases the efficacy of our proposed algorithm against real and spoof adversaries, achieving high physical similarity scores in FR models and state-of-the-art commercial systems. On average, there is only a 14\% reduction in scores from digital to physical attacks, with high attack success rate in both white- and black-box scenarios.
\end{abstract}


\section{Introduction}
Deep-learning systems have been shown to be vulnerable to adversarial examples \cite{szegedy2014intriguing,goodfellow2014explaining}, which mislead the models by adding manipulated patterns to the inputs. Face recognition (FR) systems, based on deep neural networks (DNNs), also face the threat of adversarial examples. The security issues are especially  important for these systems since they are usually deployed on critical applications such as payment, phone unlocking, access control etc. 

Most work on adversarial examples have focused on digital attacks \cite{madry2018towards,dong2018boosting,xie2019improving,dong2019evading,lin2020nesterov,xiang2021improving}, where direct accessibility to the input of DNNs is assumed. However, this assumption does not hold for a real FR system, where a photo is taken for a user and sent for inference by a backend model. Efforts have also been made to extend the notion of adversarial attack to physical domain, where the object or physical environment is manipulated so that the captured images can mislead the DNNs \cite{chen2018shapeshifter,song2018physical,thys2019fooling}. For example, 2D adversarial patches \cite{komkov2021advhat,sharif2016accessorize} are attached to physical faces to attack FR systems. A makeup generation method  was used to synthesize eye shadow over the orbital region on faces \cite{yin2021advmakeup}. However, these methods require careful calibration and fabrication of the physical artifacts \cite{liu2022rstam}.

Another flow of work modulates the light for illuminating objects \cite{sayles2021invisible,duan2021laser}. While providing enhanced flexibility, these techniques require careful programming of light-emitting devices, which are sensitive to changes in scene lighting and can only achieve non-targeted attacks. Recently, an adversarial light attack was introduced using a projector-camera setup \cite{nguyen2020light, gnanasambandam2021optical}, where adversarial patterns are digitally created and projected onto specific objects. One study \cite{nguyen2020light} suggests a global correction method to handle color shifts from projector to camera. However, this approach has limitations as it overlooks spatial variations in optical systems, limiting its applicability. Another study \cite{gnanasambandam2021optical} incorporates the projector-camera model into optimizing adversarial attacks by detailed modeling of instruments and environmental factors. Nonetheless, this method requires precise calibration and accurate pixel mapping during modeling.


In this paper, we consider a practical setup for adversarial light attacks on FR systems, utilizing a portable projector for easy deployment. We assume that consumer-grade cameras are employed by the FR systems, introducing unique challenges when applying the methodologies outlined in previous works  \cite{nguyen2020light, gnanasambandam2021optical} within this context. The accurate color calibration of the projector-camera system proves unattainable due to the low quality of the projected image.  

To address the challenges in a portable project-camera setup for attacking an FR system, we introduce a series of device-aware adaptations within the adversarial attack algorithm to mitigate losses incurred during the transition from the digital to physical domain. It is important to emphasize that our adaptation scheme seamlessly integrates with any digital attack algorithm. Our main contributions are summarized as follows:

{\noindent\bf Resolution-aware adaptation}. Acknowledging the discrepancy between the actual physical resolution of the projector and that of the digital screen, we propose a patch-based similarity loss mechanism to enforce pixel values within a patch to exhibit similarity, aiming to minimize the loss of adversarial patterns post-projection.

{\noindent\bf Color-aware adaptation}. Initially, we derive a coarse estimation of the projector's color mapping function, following the methodology outlined in \cite{gnanasambandam2021optical}. Subsequently, we dynamically adjust the valid projector color interval, simulate moiré noise, and operate in grayscale during attack optimization to mitigate color loss post-projection.

{\noindent\bf Experimental validation}. We conducted extensive experiments to validate the proposed algorithm, involving real and spoof adversaries made of various materials. Our devised scheme consistently achieves superior physical cosine similarity scores across most scenarios. On average, there is only a 14\% decrease in scores from digital to physical attacks.
\section{THE PROPOSED METHOD}
\subsection{Problem Setup}
\begin{figure}[!t]
\includegraphics[width=.485\textwidth]{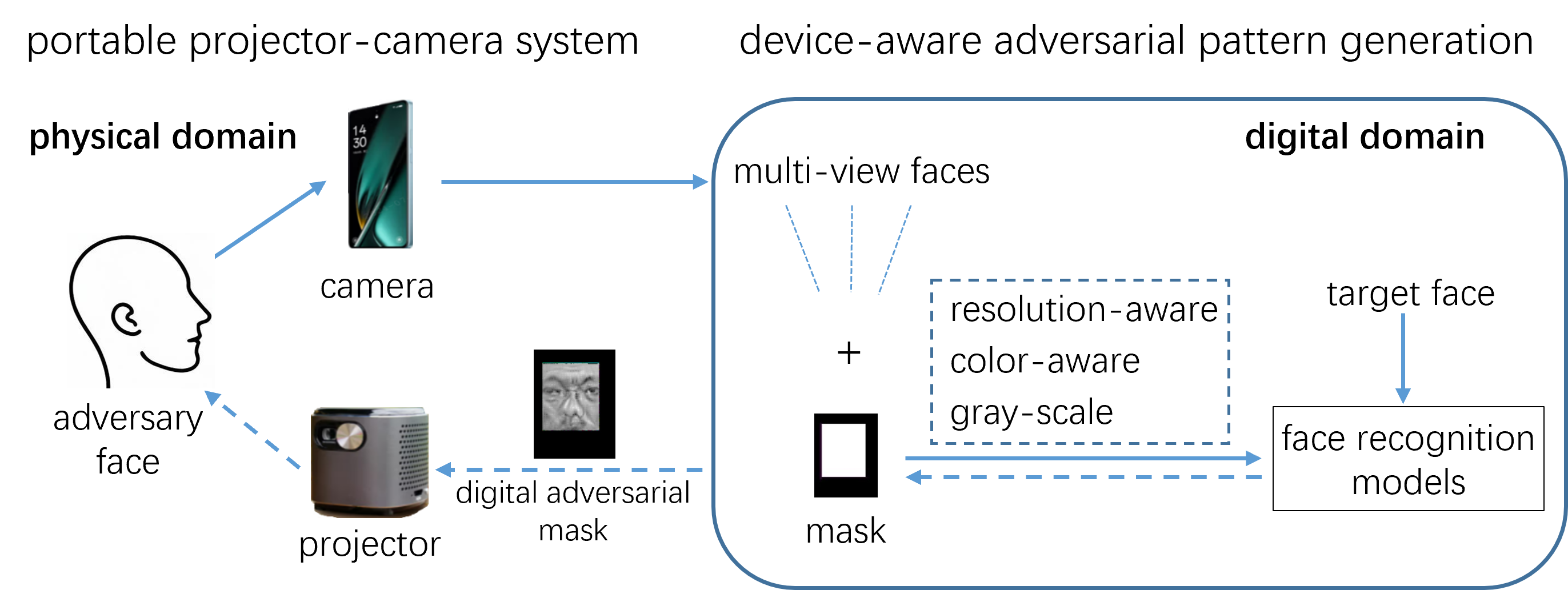} 
\caption{The overall framework.}
\vspace*{-.2in}
\label{fig:framework}
\end{figure}

As shown in Fig.~\ref{fig:framework}, our work is based on a portable projector-camera setup for attacking a face recognition system. An adversary, projected with digitally generated adversarial mask on its face, aims to prompt the FR application to recognize it as a target identity. We consider this type of impersonation attack, as it presents a more practical threat in real scenarios. As previously mentioned, several challenges exist in this setup that need to be addressed.

{\noindent\bf Calibration from digital image to physical face}. Achieving precise alignment of pixel positions is essential when projecting the adversarial digital mask onto the adversary's face to maintain adversarial effectiveness. This task is not trivial due to the different positions of the projector and camera, as well as the 2D to 3D transformation involved.

{\noindent\bf Resolution disparity}. Typically, the physical resolution of a portable projector is lower than that of the input signal, resulting in compression of the digital pattern upon projection.

\begin{figure}[!t]
\centering
\includegraphics[width=.245\textwidth]{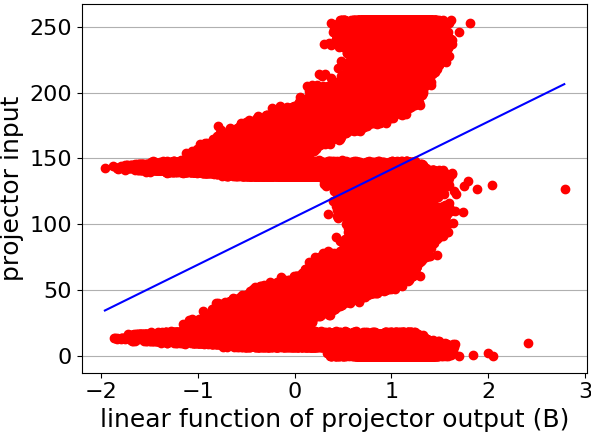} 
\vspace*{-.1in}
\caption{Rough estimation of projector radiometric response for channel B.}
\vspace*{-.2in}
\label{fig:projectorcolor}
\end{figure}

{\noindent\bf Color degradation}. In the projector-camera configuration, color transformation occurs at both the projector and the camera. Regarding the projector, the nonlinearity of the projector will alter the intensity of the input signal. Following the work in \cite{gnanasambandam2021optical,nayar2003projection}, we aim to approximate the radiometric response function of the projector, which determines the projected value for each pixel of the input signal per color channel. However, as shown in Fig.~\ref{fig:projectorcolor}, the mapping is significantly overlapped due to pronounced moiré noise and the utilization of low-quality optics.

Regarding the camera, we use a mobile phone camera as a substitute for the one utilized in the FR system. As shown in Fig. \ref{fig:spoofattack}, significant moiré noise can be observed in the captured image, resulting from disparities in the optical characteristics of the projector, the camera’s shutter speed and the refresh rate of the projected image.
\subsection{Device-aware Adversarial Pattern Generation}
In this section, we propose a device-aware optical adversarial attack algorithm, as shown in Fig. \ref{fig:framework}. Instead of utilizing an average representative image generated by applying various transformations to the adversary's face image, as done in [18], we choose to capture a brief video of the adversary with slight camera movements and select $k$ photos from different viewpoints. The faces from the photos are then cropped and aligned. A mask image is manually crafted to constrain the region in the aligned image where the adversarial pattern can be positioned. Throughout the attack, we generate a universal adversarial mask for these multi-view images to account for potential positional discrepancies between the digital and physical faces. In this way, we avoid the intricate calibration of the projector-camera setup and enable a high degree of flexibility in the physical attack procedure.

Let $x_i$ denote the $i$-th captured face of the adversary for $1 \leq i \leq k$, $y$ denote the face of the target identity, $f(x)$ denote an FR model that extracts a normalized feature representation for the input $x$, and $M$ denote the binary mask. A universal adversarial mask $x^{adv}$ is generated such that each adversary’s face projected with the mask is denoted by $x_i^\prime = x_i \odot (1-M) + x^{adv} \odot M$, where $x^{adv} \odot (1-M) = 0$ and $\odot$ represents the element-wise product. Our goal is to optimize the following:
\begin{equation*}
\vspace{-.05in}
  \mathop{argmin}_{x^{adv}} \mathcal{L}(x^{adv}, x_i^\prime, y), \text{s.t.}\ \| (x^{adv} - x_i) \odot M\|_1 \leq \epsilon 
\end{equation*}
for all $1 \leq i \leq k$ and the pixel values in the masked region of $x^{adv}$ are restricted within the $l_1$-bound of $x_i$. Here, $\epsilon$ is the perturbation bound. Essentially, $\mathcal{L}$ can be the sum of multiple cosine similarity functions, each given by $\ell^{cos}_i = 1 - \cos{\left(f(x_i^\prime), f(y)\right)}$. $x_i^\prime$ and $y$ are considered the same identity when $\ell^{cos}_i$ is above a certain threshold. Alg.~\ref{alg:attack} outlines our proposed adaptation built upon a baseline algorithm framework.
{\small
\begin{algorithm}[!t]
\caption{Device-aware Adversarial Mask Generation}
\begin{algorithmic}[1]
\Require $k$ views of adversary $x_i$, target face $y$, binary mask $M$, FR model $f$, diverse input transformation $\mathcal{T}_i$.
\Require perturbation bound $\epsilon$, decay factor $\mu$, number of iterations $N$, brightness term $\gamma$, transparency coefficient $\theta$, Gaussian kernel $G$.
\Ensure universal adversarial mask $x_N^{adv}$.
\State $\alpha=1.5\epsilon/N$, $g_0=0$
\State $x_0^{adv}={\bf c2g}(y)$
\For{$t=0$ to $N-1$}
    \State $\hat{g}_s=0$
    \For{$j=0$ to $m-1$} \Comment{SIM}
        \State generate a random moiré pattern image $\psi$
        \State ${\bf x_t^{adv-moire}}={\bf g2c}\left(S_j\left(x_t^{adv}\right)\right)\cdot \left(1-\theta\right)+\psi \cdot\theta$
        \State $x_t^{adv-m-b}=x_t^{adv-moire}\cdot(1+\gamma)$
        \State $x_i^\prime=\mathcal{T}_i(x_i\odot(1-M)+x_t^{adv-m-b}\odot M)$ \Comment{DIM}
        \State $\hat{g}_s=\hat{g}_s+\nabla_{S_j(x_t^{adv})}{\bf\mathcal{L}}\left(S_j(x_t^{adv}),x_i^\prime,y\right), \forall i$
    \EndFor
    \State $g=G*\hat{g}_s$  \Comment{TIM}
    \State $g_{t+1} = \mu \cdot g_t + \frac{g}{\|g\|_1}$ \Comment{momentum}
    \State $\hat{\mathbf{x}}_{t+1}^{\text{adv}} = \text{Clip}[x_i-\epsilon, x_i+\epsilon](x_{t}^{\text{adv}} - \alpha \cdot \frac{g_{t+1}}{\|g_{t+1}\|_1}), \forall i$
    \State $\mathbf{x}_{t+1}^{\text{adv}} = \text{{\bf color projection} of } \hat{\mathbf{x}}_{t+1}^{\text{adv}} \text{ by Eq. }(\ref{eq:3})$
\EndFor
\State \textbf{return} $\mathbf{x}_{N}^{\text{adv}}$
\end{algorithmic}
\label{alg:attack}
\end{algorithm}
}

{\noindent\bf Baseline}. We consider a representative black-box attack algorithm as the baseline. It incorporates techniques to enhance attack transferability, which include integrating the momentum into the iteration (step 13) and utilizing input transformations such as translation-invariant (TIM, step 12), diverse input method (DIM, step 9, applying transformation $\mathcal{T}_i$) and scale-invariant methods (SIM, block of step 5, where $S_j\left(x\right)$ denotes the image obtained after scaling each pixel value of $\mathit{x}$ by $\frac{1}{2^j}$. Additionally, a brightness term $\gamma$ is sampled from a uniform distribution during each iteration to account for slight environmental illumination variance. We extend the loss of the baseline algorithm with a smoothness loss $\ell_{\text{smooth}}$ \cite{duan2020adversarial}, which aims to enhance adversarial robustness by reducing the variance between adjacent pixels.

{\noindent\bf Resolution-aware}. Taking into account the resolution disparity between the physical resolution of the projector and the digital input, we introduce a patch-based loss constraint to ensure that pixel values within a patch are as similar as possible, thereby minimizing information loss after projection. We divide the adversarial mask $x^{adv}$ into $P$ patches, each of size $H\times W$, where the values of $H$ and $W$ are computed to ensure that one patch covers at least one pixel after projection. By aligning all pixel values within a patch to a one-dimensional vector $\mathbf{\hat{x}}_b^{adv}$, the patch-based loss constraint is introduced:
\vspace{-.13in}
\begin{equation}
\vspace{-.1in}
\ell_p = \frac{1}{P(HW-1)}\sum_{b=1}^{P}\sum_{j=1}^{HW-1}\left|\mathbf{\hat{x}}_b^{adv}(j)-\mathbf{\hat{x}}_b^{adv}(j-1)\right|
\end{equation}
Additionally, the smooth loss reduces the contrast at the border pixels between adjacent patches, aiding in preserving color information. The overall loss function $\mathcal{L}$ is therefore the sum of three components: $\mathcal{L}=\sum_{i=1}^k\ell^{cos}_i+\ell_{smooth}+\ell_p$.

{\noindent\bf Color-aware}. To mitigate color degradation in the portable projector-camera setup, we initially operate the input adversarial mask in grayscale ({\bf c2g}) to minimize the impact of color conversion (step 2). Next, we simulate potential moiré patterns during the attack and merge them with the adversarial mask at random locations (step 7). Subsequently, we aim to optimize the attack to operate within a linear relationship interval of the projector color mapping, enhancing alignment between the digital and physical attacks.
Based on the rough initial estimation per color channel, we dynamically adjust the linear interval range during the attack to accommodate any inaccuracies in the initial estimation. Taking channel B as an example, if the initially coarsely estimated linear interval for projector color mapping, as depicted in Fig. \ref{fig:projectorcolor}, is denoted by $[b_L,b_H]$, we work as follows.

At iteration step $t$, the computed grayscale mask $\hat{x}_{t+1}^{adv}$ is first transformed into an RGB color image $\hat{x}_{t+1}^{adv-rgb} = \textbf{g2c}(\hat{x}_{t+1}^{adv})$. For color channel B, a value $v$ is randomly sampled from a uniform distribution $[-\tau,\tau]$ and added to  each of $b_L$ and $b_H$. The pixel values of the adversarial mask are constrained within the adjusted interval (step 15) as follows:
\vspace{-.04in}
\begin{equation}
\hat{x}_{t+1,B}^{adv-rgb-proj} = \text{Clip}_{[b_L-v, b_H+v]}(\hat{x}_{t+1,B}^{adv-rgb}) 
\label{eq:3}
\vspace{-.04in}
\end{equation}
The R and G channels undergo similar processing. The clipped color image is then converted back to a grayscale image $x_{t+1}^{adv} = \textbf{c2g}(\hat{x}_{t+1}^{adv-rgb-proj})$.

Note that our adaptation adds a marginal complexity to the original digital algorithm, which is linear with the number of image pixels.
\section{Experiments}
\subsection{Experimental Setup}
We experiment with spoof and real adversaries involving ten subjects in total to verify the effectiveness of our proposed device-aware optical adversarial attack method. Real persons are chosen as target identities and are excluded from the adversary set. Three representative FR models, including ArcFace \cite{deng2019arcface}, MobileFace \cite{chen2018mobilefacenets} and FaceNet \cite{schroff2015facenet}, are adopted for the evaluation. Each experiment is conducted in an office environment with fixed lighting, using a Huawei Nova9 phone camera and a Lenovo T6S portable projector.

{\noindent\bf Preprocessing}. A brief video is recorded for the adversary by positioning the camera in front of the face with a slight movement. Five photos are then chosen from the video to cover different viewpoints: top, middle, bottom, left, and right. A real person's image is randomly selected from the target identity set. Each of these photos undergoes face detection using the RetinaFace-based algorithm \cite{deng2020retinaface} and is aligned to the size of $896\times896$, which is chosen to closely match the resolution of the projector input ($1920\times1080$). Note that a resize layer is integrated with the FR model to adjust the input image to the expected shape, while the attack is carried out on the source image of size $896\times896$. The captured photos of the projected face also undergo the same preprocessing steps.

{\noindent\bf Hyper-parameters}. The number of iterations $N$ and perturbation bound $\epsilon$ are set to 50 and 16 respectively. A brightness term $\gamma$ is sampled from a uniform distribution of [-0.3, 0.3] during each iteration.  Based on the physical parameters of the projector used, the patch size for the resolution-aware loss constraint is set to $8 \times 8$. The linear intervals for the RGB channels of the projector color mapping are selected as [63, 242], [25, 216], and [25, 204] respectively. The dynamic range per color channel is set to [-5, 5], and the transparency coefficient $\theta$ is 0.4. The other hyper-parameters related to the baseline algorithm follow the usual settings \cite{dong2018boosting,xie2019improving,dong2019evading,lin2020nesterov}.

{\noindent\bf Evaluation Metrics}. The cosine similarity score between each adversarial image and the target face is calculated. In the digital attack scenario, the top score is recorded for the five adversarial faces generated for each adversary. In the physical attack setup, three photos are taken of the adversary's face with a projected mask, captured from different viewpoints around the projector. The top score among the three captured adversarial faces is noted. The difference in similarity scores between the digital and physical attacks is then computed to demonstrate the effectiveness of the proposed method in minimizing the reduction when transitioning from the digital to the physical domain.

At the same time, the attack success rate (ASR) can be deduced, which indicates the proportion of adversaries whose similarity scores exceed a certain threshold. We utilize the threshold values from \cite{xiao2020delving}, which yield the highest accuracy on the LFW dataset (0.28, 0.21, 0.43 for ArcFace, MobileFace, and FaceNet respectively).
\subsection{Results on Spoof Adversaries}
\begin{figure}[!t]
\centering
\includegraphics[width=.158\textwidth,height=3.2cm]{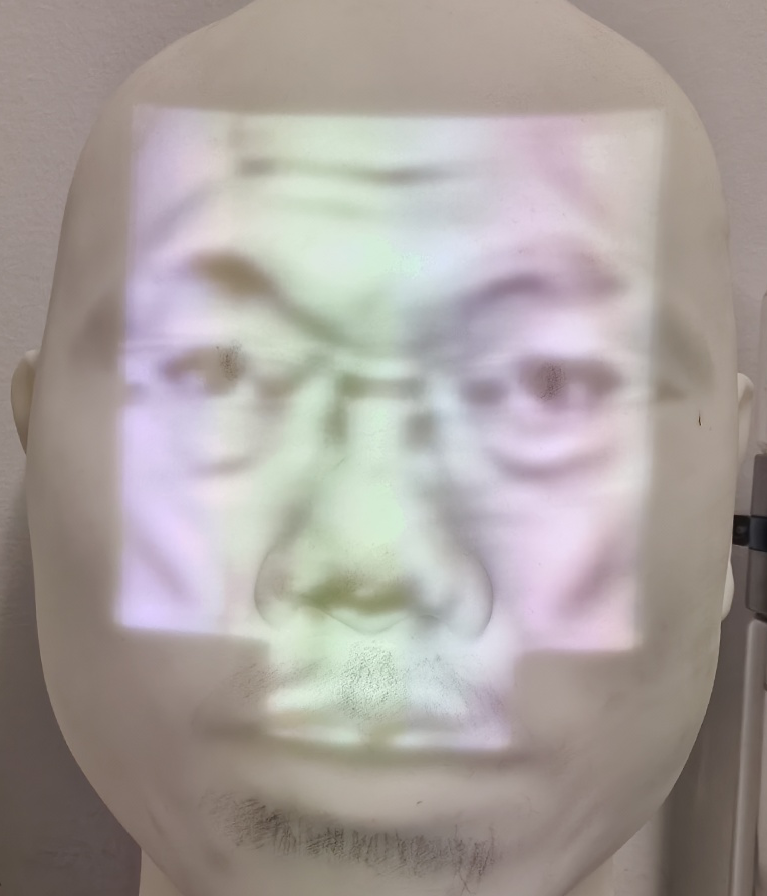} \hfill
\centering
\includegraphics[width=.158\textwidth,height=3.2cm]{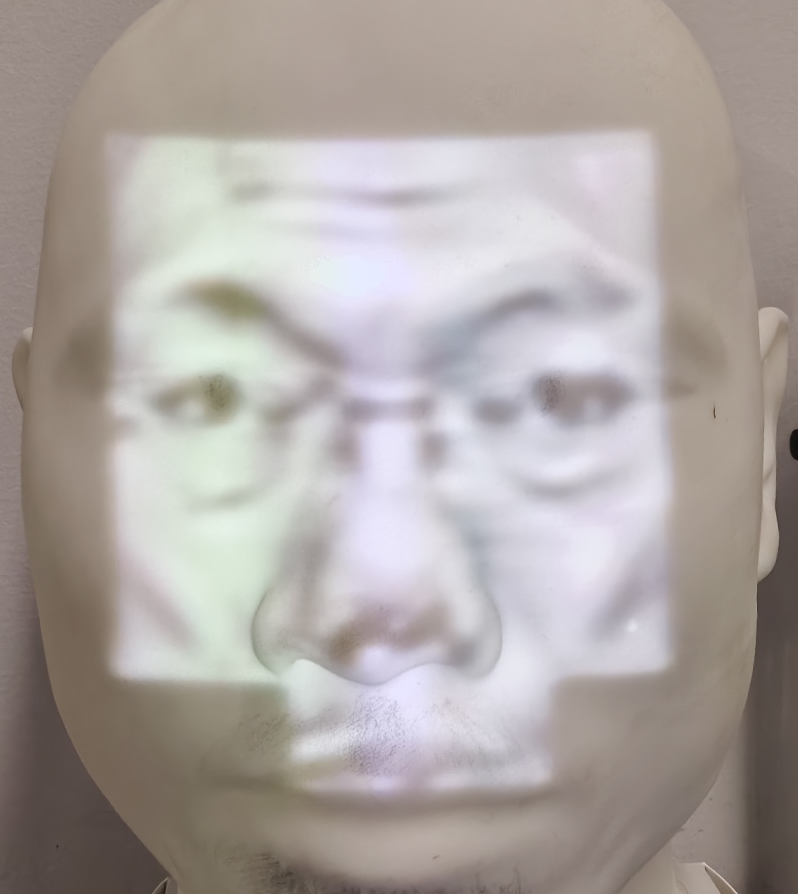} \hfill
\centering
\includegraphics[width=.158\textwidth,height=3.2cm]{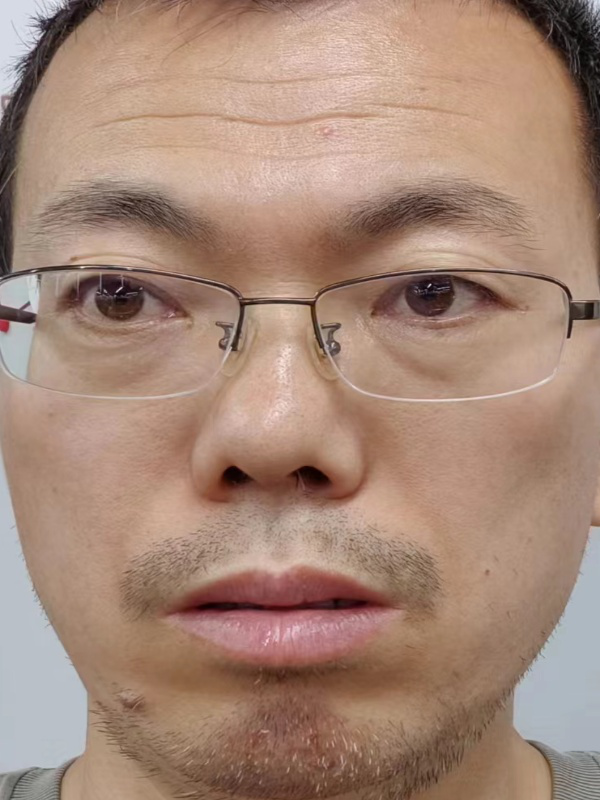} \\
\hspace{1.3cm}(a)\hfill(b)\hfill(c)\hspace{1.2cm}
\vspace*{-.1in}
\caption{Example of an impersonation attack on ArcFace in a white-box setting. Images (a) and (b) show a plastic head mode under projected adversarial light,  both recognized as the target (c).}
\vspace*{-.15in}
\label{fig:spoofattack}
\end{figure}

We experiment with five spoof adversaries with various materials, including one plastic head mold, one gypsum head mold, two resin masks, and one silicone mask. In each run, we select one of the three FR models as the substitute model to conduct a white-box attack, while the other two are used as black-box models. A universal adversarial mask is generated for all five photos of each adversary. In the digital attack, each adversary’s face, masked with the adversarial pattern, is evaluated with the randomly selected target face. In the physical attack, the adversarial mask is projected onto the adversary’s face and then captured by a camera from three viewpoints, which are used to evaluate with the target face. Fig. \ref{fig:spoofattack} shows an example of the physical attack with a spoof adversary. 

TABLE \ref{tab:spoofscore} shows the average cosine similarity score for all five spoof adversaries for each of the FR models. The upper part displays the results for gray-scale based attacks, while the lower part shows the results for color-scale based attacks. The bold and underlined entries indicate the better results between gray and color scale based attacks. It can be observed that in most cases, the generated adversarial masks in gray scale have higher similarity scores and lower reduction loss after projection.
\subsection{Results on Real Adversaries}
We also experiment with five real persons, following the same procedure. As shown in TABLE \ref{tab:realscore}, similar results can be observed as with the spoof adversaries that adversarial projections in grayscale outperform those in color. If the top score for a physical projection exceeds the threshold, the attack is deemed successful \cite{nguyen2020light}. We found that the ASR for all the spoof and real physical adversaries achieves nearly 100\%.
\subsection{Ablation Experiments}
We conduct ablation experiments with a white plastic head mold. With a randomly selected target face, we capture five photos of the mode and apply the attack algorithm on ArcFace to generate a universal adversarial mask for all source faces. Three photos are taken for the physical adversary with the projected adversarial mask. The top similarity score is recorded for both the digital and physical attacks. As shown in TABLE \ref{tab:abl} , resolution-aware adaption yields the best for the digital attack, while color-aware adaption reduces the loss in color conversion and achieves the highest physical scores in the physical domain.
\begin{table}[!t]
    \centering
    \caption{Similarity Scores for Spoof Adversaries }
    \vspace{-.05in}
    \setlength\tabcolsep{3.95pt}
    \begin{tabular}{c|ccc|ccc|ccc}
        \hline
        \textbf{Model} & \multicolumn{3}{c|}{\textbf{ArcFace}} & \multicolumn{3}{c|}{\textbf{MobileFace}} & \multicolumn{3}{c}{\textbf{FaceNet}} \\
        \cline{2-10}
         & \textbf{Dig.} & \textbf{Phy.} & \textbf{Diff.} & \textbf{Dig.} & \textbf{Phy.} & \textbf{Diff.} & \textbf{Dig.} & \textbf{Phy.} & \textbf{Diff.} \\
        \hline
        \textbf{ArcFace} & 0.80 & \textbf{0.63} & \underline{0.17} & 0.61 & \textbf{0.54} &  \underline{0.07} & 0.79 & \textbf{0.75} &  \underline{0.04} \\
        \textbf{MobileFace} & 0.64 & \textbf{0.52} & 0.12 & 0.78 & 0.54 & 0.24 & 0.69 & \textbf{0.68} &  \underline{0.01} \\
        \textbf{FaceNet} & 0.50 & \textbf{0.44} &  \underline{0.06} & 0.42 & \textbf{0.38} &  \underline{0.04} & 0.92 & \textbf{0.78} &  \underline{0.14} \\
        \hline
        \hline
        \textbf{ArcFace} & 0.84 & 0.63 & 0.20 & 0.59 & 0.50 & 0.10 & 0.79 & 0.75 & 0.05 \\
        \textbf{MobileFace} & 0.62 & 0.51 &  \underline{0.11} & 0.77 & \textbf{0.54} &  \underline{0.23} & 0.69 & 0.68 & 0.06 \\
        \textbf{FaceNet} & 0.51 & 0.39 & 0.12 & 0.41 & 0.35 & 0.06 & 0.92 & 0.78 & 0.15 \\
        \hline
    \end{tabular}
    {\noindent{Note: upper part for gray-scale and bottom part for color-scale.}}
\label{tab:spoofscore}
\vspace{-.1in}
\end{table}
\begin{table}[!t]
    \centering
    \caption{Similarity Scores for Real Adversaries}
    \vspace{-.05in}
    \setlength\tabcolsep{3.95pt}
    \begin{tabular}{c|ccc|ccc|ccc}
        \hline
        \textbf{Model} & \multicolumn{3}{c|}{\bf ArcFace} & \multicolumn{3}{c|}{\bf MobileFace} & \multicolumn{3}{c}{\bf FaceNet} \\
         \cline{2-10}
         & \textbf{Dig.} & \textbf{Phy.} & \textbf{Diff.} & \textbf{Dig.} & \textbf{Phy.} & \textbf{Diff.} & \textbf{Dig.} & \textbf{Phy.} & \textbf{Diff.} \\
        \hline
        \textbf{ArcFace} & 0.77 & \textbf{0.67} & \underline{0.10} & 0.61 & \textbf{0.56} & \underline{0.05} & 0.77 & \textbf{0.71} & \underline{0.06} \\
        \textbf{MobileFace} & 0.55 & \textbf{0.47} & \underline{0.08} & 0.72 & \textbf{0.52} & 0.20 & 0.69 & \textbf{0.62} & \underline{0.06} \\
        \textbf{FaceNet} & 0.49 & \textbf{0.46} & \underline{0.03} & 0.42 & 0.31 & \underline{0.11} & 0.88 & \textbf{0.68} & \underline{0.21} \\
        \hline
        \hline
        \textbf{ArcFace} & 0.75 & 0.64 & 0.11 & 0.58 & 0.50 & 0.08 & 0.75 & 0.66 & 0.09 \\
        \textbf{MobileFace} & 0.57 & 0.44 & 0.13 & 0.66 & 0.47 & \underline{0.19} & 0.68 & 0.54 & 0.15 \\
        \textbf{FaceNet} & 0.51 & 0.44 & 0.07 & 0.44 & \textbf{0.32} & 0.12 & 0.85 & 0.61 & 0.24 \\
        \hline
    \end{tabular}
    {\noindent{Note: upper part for gray-scale and bottom part for color-scale.}}
    \label{tab:realscore}
    \vspace{-.1in}
\end{table}
\begin{table}[!t]
    \centering
    \caption{SIM. SCORES FOR ABLATION STUDIES AT GRAY SCALE}
    \vspace{-.05in}
    \setlength\tabcolsep{3.95pt}
    \begin{tabular}{c|ccc|ccc|ccc}
        \hline
        \textbf{Model} & \multicolumn{3}{c|}{\bf ArcFace} & \multicolumn{3}{c|}{\bf MobileFace} & \multicolumn{3}{c}{\bf FaceNet} \\
        \cline{2-10}
         & \textbf{Dig.} & \textbf{Phy.} & \textbf{Diff.} & \textbf{Dig.} & \textbf{Phy.} & \textbf{Diff.} & \textbf{Dig.} & \textbf{Phy.} & \textbf{Diff.} \\
        \hline
        \textbf{baseline} & 0.85 & 0.69 & 0.16 & 0.69 & 0.61 & \underline{0.08} & 0.84 & 0.66 & 0.18 \\
        \textbf{res.-aware} & 0.86 & 0.71 & 0.15 & 0.72 & 0.60 & 0.12 & 0.85 & 0.73 & 0.12 \\
        \textbf{color-aware} & 0.86 & \textbf{0.74} & \underline{0.12} & 0.71 & \textbf{0.62} & 0.09 & 0.83 & \textbf{0.76} & \underline{0.07} \\
        \hline
    \end{tabular}
    \label{tab:abl}
    \vspace{-.1in}
\end{table}
\subsection{Commercial Systems}
We randomly select ten captured faces projected with the adversarial mask generated from the ArcFace model, with one photo per adversary, to evaluate with three commercial FR systems: Face++ \cite{faceplusplus}, Baidu \cite{baidu} and Tencent \cite{tencent}. The mean of the recorded confidence scores are around 70.2\%, 80.7\% and 45.8\% for these three systems respectively. 
\vspace{.16in}
\section{CONCLUSIONS}
\vspace{.1in}
In this paper, we investigate optical adversarial attacks on FR systems using a portable projector-camera setup. To address the loss in adversarial effectiveness resulting from the projection and recapture of digitally generated patterns, we introduce device-aware adaptation during the attack optimization process. Specifically, we enforce similarity constraints on pixel values within patches to mitigate color compression caused by the disparity between physical and digital screen resolutions in portable projectors. Given the challenge of obtaining the precise radiometric response function of the projector, we project pixel values within a dynamically adjusted valid linear interval during each iteration. Our experiments demonstrate that the proposed adaptation algorithm in grayscale yields the most effective results for physical attacks.
\onecolumn
\newpage
\twocolumn
\bibliographystyle{IEEEtran}
 \bibliography{IEEEabrv,papers}
\end{document}